\documentclass{article}
\usepackage{ijcai26}
\usepackage{natbib}
\usepackage{times}
\usepackage{soul}
\usepackage{url}
\usepackage[hidelinks]{hyperref}
\usepackage[utf8]{inputenc}
\usepackage[small]{caption}
\usepackage{graphicx}
\usepackage{amsmath}
\usepackage{amsthm}
\usepackage{booktabs}
\usepackage{algorithm}
\usepackage[switch]{lineno}
\usepackage[table,xcdraw]{xcolor}
\usepackage{amssymb, amsfonts} 
\usepackage{multirow}
\newtheorem{lemma}{Lemma}

\usepackage{xcolor}
\usepackage{amssymb}
\usepackage{algpseudocode}
\usepackage{graphicx}

\newtheorem{theorem}{Theorem}

\title{
  \raisebox{-0.8em}{
    \includegraphics[height=2.3em]{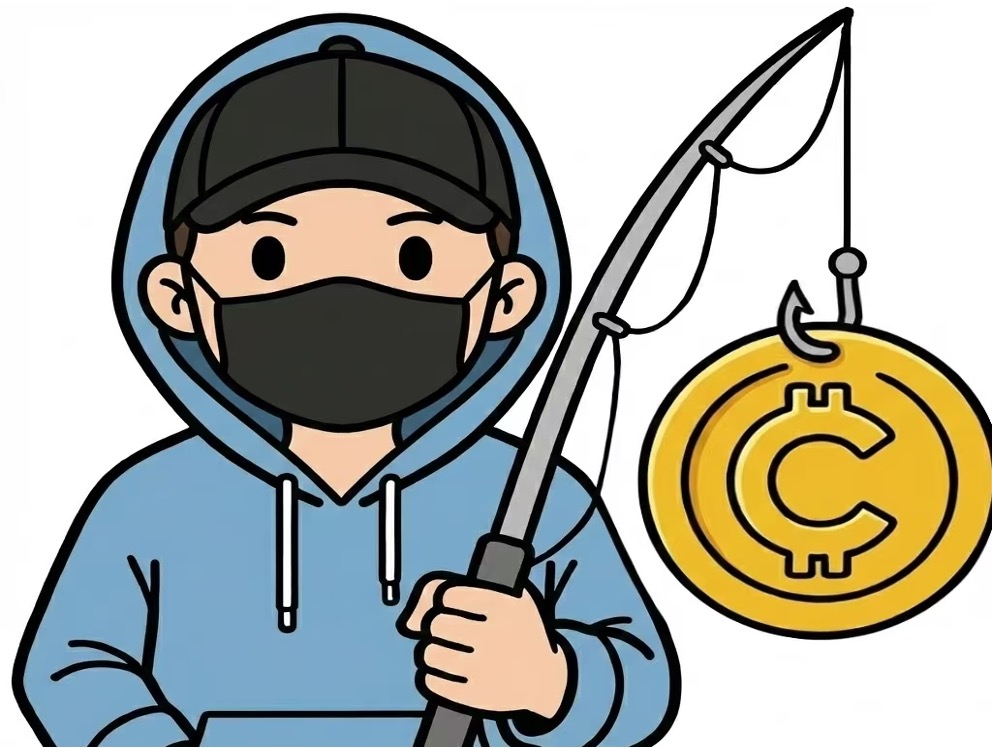}
  }\;LURE: Latent Space Unblocking for Multi-Concept Reawakening in Diffusion Models}

\author{
Mengyu Sun$^{1,2}$
\and
Ziyuan Yang$^{1,3}$\thanks{\raggedright Corresponding authors: cziyuanyang@gmail.com, junxu.liu@polyu.edu.hk} 
\and
Andrew Beng Jin Teoh$^4$
\and
Junxu Liu$^{2,}$\footnotemark[1]\\
\and
Haibo Hu$^2$
\and
Yi Zhang$^1$
\\
\affiliations
$^1$Sichuan University, China,
$^2$The Hong Kong Polytechnic University, Hong Kong SAR, China,\\
$^3$Nanyang Technological University, Singapore,
$^4$Yonsei University, South Korea
}

\begin{document}

\maketitle

\begin{abstract}
Concept erasure aims to suppress sensitive content in diffusion models, but recent studies show that erased concepts can still be reawakened, revealing vulnerabilities in erasure methods. Existing reawakening methods mainly rely on prompt-level optimization to manipulate sampling trajectories, neglecting other generative factors, which limits a comprehensive understanding of the underlying dynamics. 
In this paper, we model the generation process as an implicit function to enable a comprehensive theoretical analysis of multiple factors, including textual conditions, model parameters, and latent states. We theoretically show that perturbing each factor can reawaken erased concepts.
Building on this insight, we propose a novel concept reawakening method: \textbf{\textit{L}}atent space \textbf{\textit{U}}nblocking for concept \textbf{\textit{RE}}awakening~(\textbf{\textit{LURE}}), which reawakens erased concepts by reconstructing the latent space and guiding the sampling trajectory.
Specifically, our semantic re-binding mechanism reconstructs the latent space by aligning denoising predictions with target distributions to reestablish severed text-visual associations.
However, in multi-concept scenarios, naive reconstruction can cause gradient conflicts and feature entanglement. To address this, we introduce Gradient Field Orthogonalization, which enforces feature orthogonality to prevent mutual interference. Additionally, our Latent Semantic Identification-Guided Sampling~(LSIS) ensures stability of the reawakening process via posterior density verification. Extensive experiments demonstrate that LURE enables simultaneous, high-fidelity reawakening of multiple erased concepts across diverse erasure tasks and methods.
Code is available at: \hyperlink{https://github.com/Katherine1312/LURE}{https://github.com/Katherine1312/LURE}.
\end{abstract}

\section{Introduction}
Text-to-image diffusion models (DMs) have become a standard paradigm for image synthesis from natural-language prompts and support diverse creative applications~\citep {zhang2023text}. However, their expressive power also introduces significant risks of misuse, including the generation of harmful or unauthorized content~\citep{schramowski2023safe}. To mitigate these risks, concept-erasure techniques have been proposed to suppress or remove sensitive content, ranging from explicit content, private identities, and copyrighted material~\citep{xie2025erasing,li2025speed}.

Despite their effectiveness, recent studies reveal that erased concepts can often be restored after erasure~\citep{richardson2025rethinking}, indicating weaknesses in current erasure pipelines. Most concept-erasure methods leave the underlying latent space intact and instead suppress concept expression by blocking specific sampling trajectories~\citep{xie2025erasing}. Consequently, concept-reawakening methods seek alternative trajectories that recover suppressed concepts under modified inputs. Early approaches such as textual inversion~\citep{pham2023circumventing} and prompt optimization~\citep{zhang2024generate} leverage embedding shifts or linguistic reformulations to bypass erasure barriers.

More recent works have explored discrete token search strategies~\citep{Yang2023SneakyPromptJT} and semantic decomposition approaches~\citep{wang2024chain}, which manipulate prompt structures to circumvent erasure mechanisms. Additionally, \citet{beerens2025vulnerability} proposed employing coordinate descent to iteratively optimize prompt tokens to elicit erased concepts.

While existing works on concept erasure and recovery in DMs primarily focus on modifying sampling trajectories, they largely neglect other fundamental components that jointly govern the diffusion process. To gain a more comprehensive understanding of the latent mechanisms underlying concept erasure, we conduct a theoretical analysis by formulating diffusion-based text-to-image generation as an implicit function. Specifically, we model text-to-image generation as an implicit mapping that characterizes the alignment between textual conditions and generated visual outputs, which is jointly determined by the model parameters, the text prompt, and the latent state.

Most existing concept-erasure approaches avoid modifying the latent space, as doing so may introduce unintended interference on preserved concepts. Instead, they suppress concept expression by obstructing specific sampling trajectories. Correspondingly, current concept reawakening methods aim to circumvent such suppression by perturbing the text-conditioning component, typically through adversarial or optimized text embeddings while keeping the model parameters and latent dynamics fixed.

\begin{figure}[t]
    \centering
    \includegraphics[width=1\linewidth]{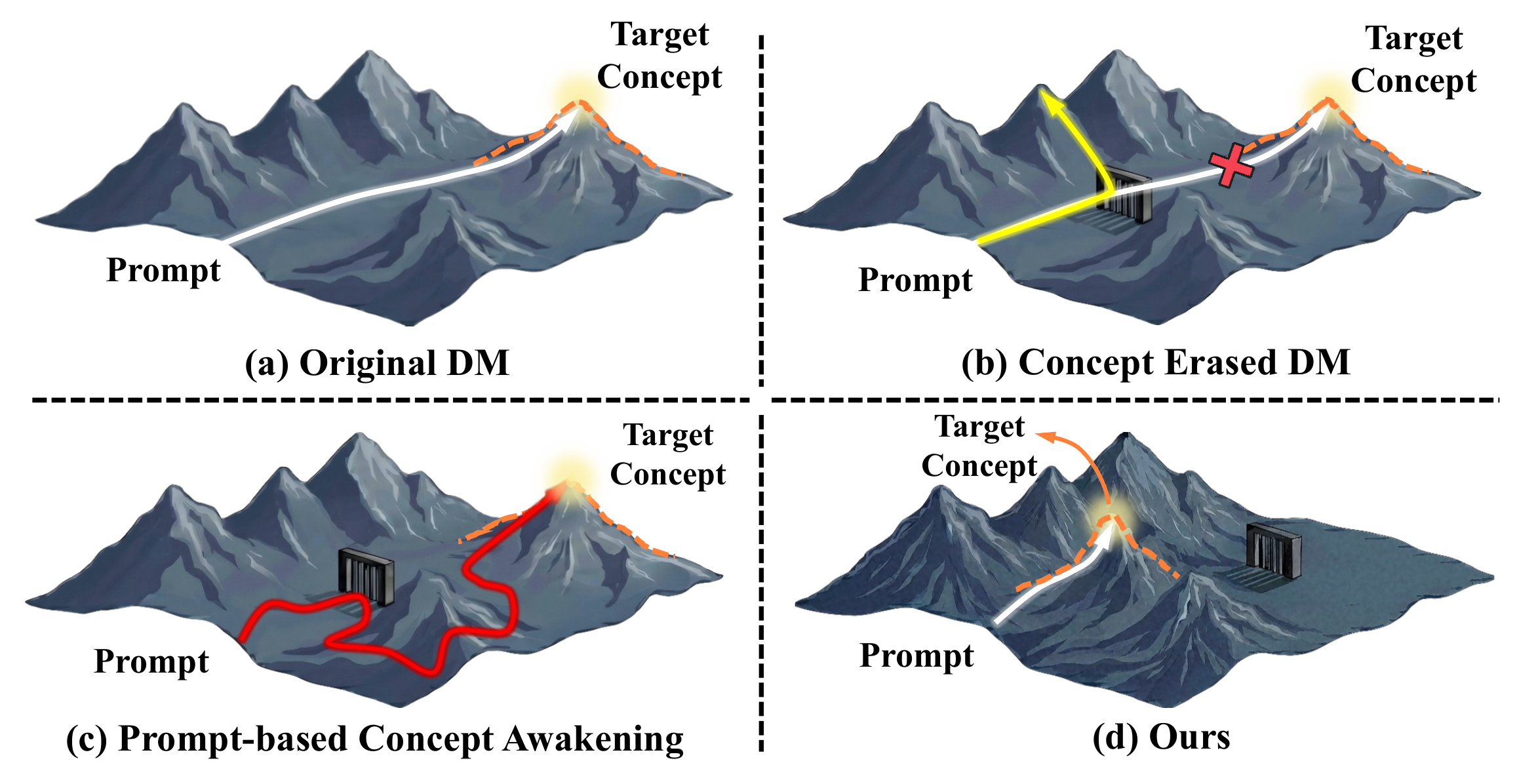}
    \vspace{-20pt}
    \caption{Concept illustrations of concept erasure and reawakening paradigms. (a) Original prompt–concept alignment enables direct semantic generation. (b) Concept erasure suppresses target concepts by blocking the original sampling trajectory. (c) Prompt-based methods seek alternative trajectories by optimizing or reformulating the prompts to bypass the erasure barrier. (d) Our method reconstructs the latent space to unblock the direct semantic trajectory.}
\vspace{-15pt}
    \label{img-intro}
\end{figure}

From the implicit-function perspective, these methods effectively constrain the output by modifying only the textual input to the implicit mapping. In contrast, if the implicit function itself is partially reconstructed, the erased concepts can be recovered along the original sampling trajectory without relying on adversarial prompt manipulation. However, the main challenge lies in \textbf{\textit{how to reconstruct the implicit mapping while avoiding concept entanglement and preventing interference with other preserved concepts.}} To facilitate intuition and clarify the overall process, we provide a conceptual illustration in Fig.~\ref{img-intro}.

To answer this question, we propose a novel \textbf{\textit{L}}atent space \textbf{\textit{U}}nblocking method for concept \textbf{\textit{RE}}awakening~(\textbf{\textit{LURE}}). Our approach is designed to simultaneously achieve two objectives. First, it induces targeted shifts in the latent representations of erased concepts, thereby enabling them to circumvent existing erasure constraints. Second, it explicitly minimizes interference with the latent representations of non-erased concepts, thereby preserving overall generation quality and semantic fidelity.

To fulfill these objectives, LURE integrates two complementary modules that reawaken erased concepts via latent-space reconstruction and sampling-level guidance, respectively. First, we implement a semantic re-binding mechanism that reconstructs the latent space by enforcing the model's denoising predictions to align with the latent distribution of the target concept. In multiple-concept reawakening scenarios, naive joint optimization can lead to gradient conflicts, where the optimization signals from different concepts interfere with one another. To alleviate this issue, we introduce a novel loss that encourages the latent embeddings of different concepts to be mutually orthogonal, reducing feature entanglement and mitigating cross-concept interference. Second, we develop a Latent Semantic Identification-guided Sampling (LSIS) strategy during inference, which validates whether generated samples lie within the high-density region of the target concept’s posterior distribution. This posterior checking mechanism prevents semantic drift and stabilizes the reawakening process. By jointly leveraging these two modules, LURE enables the simultaneous, high-fidelity recovery of multiple erased concepts using a single model tuning, while preserving the overall quality and consistency of generated images. Our main contributions are summarized below:

\begin{itemize}
\item We propose a novel latent-space unblocking paradigm for concept reawakening and provide an implicit-function–based theoretical perspective on the latent mechanisms underlying concept erasure.
\item We introduce a latent-space reconstruction framework that explicitly addresses feature entanglement in multi-concept reawakening, enabling simultaneous and high-fidelity recovery of multiple concepts.
\item We propose a latent-semantic identification-guided sampling strategy that performs posterior verification during inference to mitigate semantic drift and stabilize concept reawakening.
\end{itemize}

\section{Related Works}
\noindent\textbf{Concept Erasure.}
Concept erasure technologies aim to prevent DMs from generating visual content associated with specific textual concepts. A representative example is Erased Stable Diffusion (ESD)~\citep{gandikota2023erasing}, which introduces a fine-tuning strategy based on classifier-free guidance to mitigate the generation of undesired concepts. To improve efficiency, several closed-form editing methods are proposed. TIME~\citep{orgad2023editing} and UCE~\citep{gandikota2024unified} modify cross-attention parameters using analytical solutions derived from Tikhonov regularization to achieve one-step concept erasure without iterative optimization. Building on this line of work, \citet{lu2024mace} achieve scalable mass concept erasure by combining closed-form solutions with LoRA adapters. RECE~\citep{gong2024reliable} further improves efficiency, which accomplishes reliable erasure within seconds through efficient cross-attention matrix modifications and iterative embedding derivation. SPEED~\citep{li2025speed} enhances erasure precision by enforcing null-space constrained parameter editing, supplemented by prior knowledge refinement strategies. Beyond single-shot erasure, \citet{zhang2024forget} propose a continual learning framework to preserve overall model utility during concept removal. More recent efforts emphasize parameter efficiency and localization. ConAda~\citep{lyu2024one} introduces a one-dimensional adapter-based manipulation method, while ~\citet{lee2025localized} enable spatially localized concept erasure within specific image regions using gated low-rank adaptation. Moreover, growing attention has been devoted to multi-concept erasure~\citep{liu2025dyme}. 

\noindent\textbf{Concept Reawakening.}
Alongside the development of concept-erasure techniques, prior work has investigated robustness evaluation using adversarial embedding-based attacks. For example, textual inversion–based methods~\citep{pham2023circumventing} and gradient-based prompt optimization methods~\citep{zhang2024generate} recover erased concepts by optimizing continuous text embedding vectors to reawaken suppressed semantic content.
To improve scalability and effectiveness, discrete token optimization methods have also been developed. SneakyPrompt~\citep{Yang2023SneakyPromptJT} and Ring-A-Bell~\citep{hsu2024ring} search over the discrete vocabulary space to identify adversarial token sequences that evade erasure constraints while preserving semantic coherence. More recently, semantic jailbreaking strategies~\citep{zhang2025metaphor} have been explored, which manipulate prompts by decomposing prohibited concepts into benign attributes or employing metaphorical substitutions to circumvent erasure mechanisms.
Collectively, these methods demonstrate that current erasure techniques primarily disrupt explicit text-visual bindings while leaving the underlying generative knowledge largely intact.
Complementary to prompt-level manipulation, we explore concept recovery via latent-space reconstruction, in which parameter-level tuning induces targeted latent transformations that restore severed semantic associations. This perspective enables systematic and simultaneous reawakening of multiple erased concepts, which extends beyond the limitations of prompt-based attacks.

\begin{figure*}[ht]
    \centering
    \includegraphics[width=.95\linewidth]{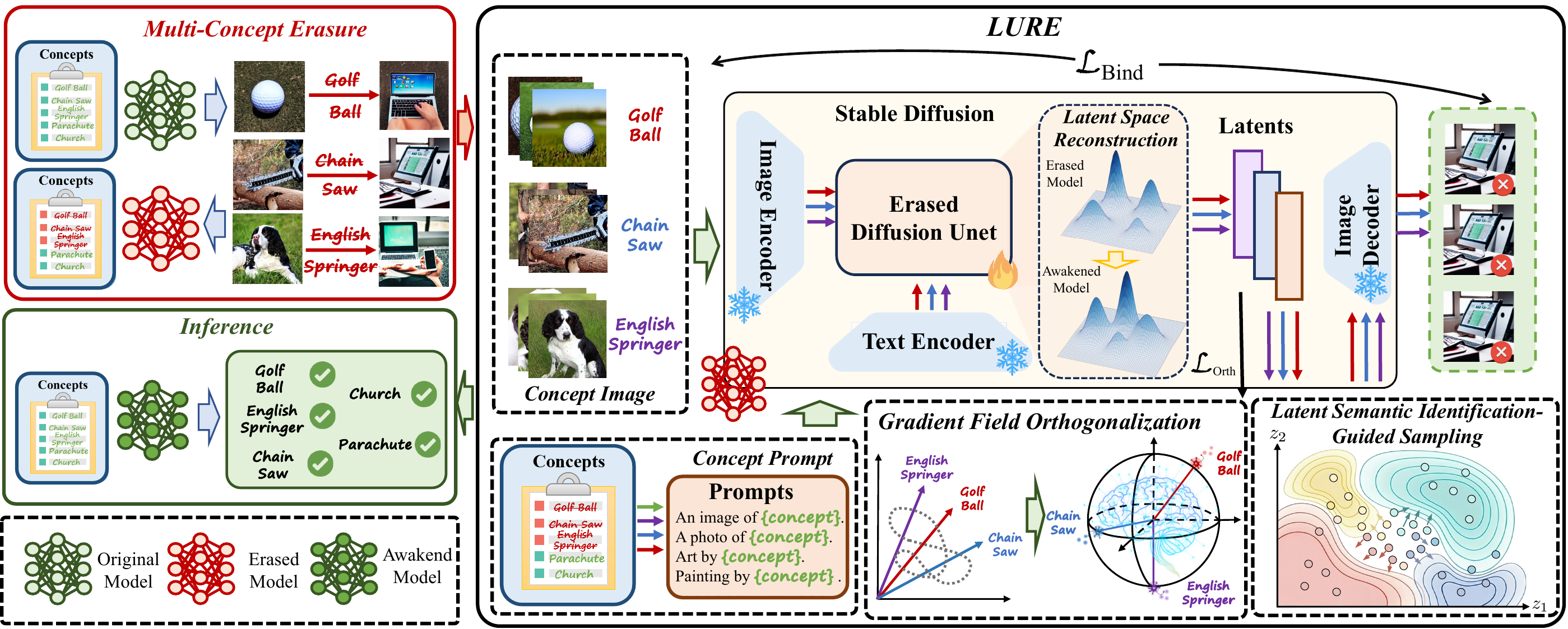}
    \vspace{-10pt}
    \caption{Illustration of the proposed LURE framework, which consists of two core components: the latent space reconstruction and latent semantic identification-guided sampling.}
    \vspace{-15pt}
    \label{fig:enter-label}
\end{figure*}

\newtheorem{definition}{Definition}
\newtheorem{assumption}{Assumption}

\section{Theoretical Analysis}
\subsection{Preliminaries and Problem Formulation}
Let $\epsilon_\theta$ denote the noise prediction network of a pre-trained DM, parameterized by $\theta$. We consider the concept set as $\mathcal{C} = \{c_1, c_2, \dots, c_D\}$, where $D$ is the number of concepts. Each concept $c_i$ corresponds to a semantic attribute that can be expressed in the generated images through its associated textual conditioning $y_{c_i}$.

\begin{definition}[\textbf{Text-Visual Alignment Function}]
We define an alignment function $\mathcal{F}$ to quantify the semantic consistency between a textual condition $y_c$ and the latent representation generated by the DM. The function is formulated as:
\label{definition 1}
\begin{equation}
\mathcal{F}(y_c, \theta, z)= -\mathbb{E}_{t, \epsilon} \left[ \log p_\phi(y_c \mid \epsilon_\theta(z_t, t, y_c)) \right],
\label{eq:1}
\end{equation}
where $p_\phi\big(y_c\mid\epsilon_\theta(\cdot)\big)$ denotes the conditional probability of observing $y_c$ given the predicted noise from $\epsilon_\theta$. $z_t = \sqrt{\bar{\alpha}_t}{z_0} + \sqrt{1 - \bar{\alpha}_t}\epsilon$ is the noisy latent at timestep $t$, with $\epsilon \sim \mathcal{N}(0, \mathbf{I})$ representing the standard Gaussian noise. The term $z_0=\omega(x_0)$ denotes the clean latent encoded from the input image $x_0$ by the visual encoder $\omega$, and $\bar{\alpha}_t$ represents the noise schedule coefficients.
\end{definition}

\begin{assumption}[\textbf{Smoothness}]
\label{Assumption 1}
The network $\epsilon_\theta(z_t, t, y)$ is continuously differentiable with respect to parameter $\theta$.
\end{assumption}

\subsection{Concept Entanglement Analysis}
\label{Concept Entanglement Theory}
\begin{theorem}[\textbf{Concept Entanglement}]
\label{theorem Concept Entanglement Theory}
We define the reconstruction loss for concept $c_i$ as:
\begin{equation}
\small
\mathcal{L}_i = \mathbb{E}_{t, x_0, \epsilon}
\left[
\left\|
\epsilon
-
\epsilon_\theta
\!\left(
\sqrt{\bar{\alpha}_t}\, z_0
+
\sqrt{1 - \bar{\alpha}_t}\, \epsilon,
\; t,
\; y_{c_i}
\right)
\right\|^2
\right],
\end{equation}
where $x_0$ is sampled from the data distribution $p(x|c_i)$ for $c_i$.

Then, let $c_i, c_j \in \mathcal{C}$ be two distinct concepts with $i \neq j$. The gradients of the two concepts exhibit non-orthogonality:
\begin{equation}
    \langle \nabla_\theta \mathcal{L}_i, \nabla_\theta \mathcal{L}_j \rangle \neq 0.
\end{equation}

Different concepts may become entangled in the latent space in the absence of the lack of explicit constraints during training. This statement should be interpreted as a general observation rather than a strict worst-case guarantee.
\end{theorem}

\begin{lemma}[\textbf{Parameter Sharing}]
\label{Parameter Sharing}
All concepts share the same query, key, and value projection matrices $\{\mathbf{W}_Q,\ \mathbf{W}_K,\ \mathbf{W}_V\}$ in the cross-attention layers, as well as other parameters of convolutional layers.
\end{lemma}

\begin{theorem}[\textbf{Gradient Conflict}]
\label{Gradient-Conflict}
For two distinct concepts $c_i$ and $c_j$ with distinct visual semantics, we define the gradient alignment coefficient $\rho_{i,j}$ as
\begin{equation}
    \rho_{i,j} = \frac{\langle \nabla_\theta \mathcal{L}_i, \nabla_\theta \mathcal{L}_j \rangle}{\|\nabla_\theta \mathcal{L}_i\| \cdot \|\nabla_\theta \mathcal{L}_j\|}.
\end{equation}

The coefficient $\rho_{i,j}$ measures the alignment degree between the optimization directions induced by the two concepts. When the optimal parameter configurations corresponding to $c_i$ and $c_j$ are not identical, their gradients are generally neither orthogonal nor fully aligned, yielding $0 < \rho_{i,j} < 1$. Under joint optimization, the resulting parameters $\theta^\ast$ therefore represent a compromise between the competing gradient directions, which may deviate from the concept-specific optima obtained when optimizing $\mathcal{L}_i$ and $\mathcal{L}_j$ independently.
\end{theorem}

\subsection{Concept Erasure}
\begin{theorem}[\textbf{Erasure-Induced Shift}]
\label{thm:erasure}
Let $\theta_0$ and $\theta_e$ denote the original and erased model parameters, respectively. The erasure operation is formulated as:
$\theta_e = \arg\min_{\theta} \mathbb{E}_{x \sim p(x|c)}
\left[
\|\epsilon_\theta(z_t, t, y_c) - \epsilon_\theta(z_t, t, y_{\emptyset})\|_2^2
\right]
+ \gamma \|\theta - \theta_0\|_2^2$,
where $y_{\emptyset}$ denotes an empty or null text embedding, $y_c$ denotes the text embedding of concept $c$ and $\gamma$ is a regularization coefficient.
\end{theorem}

\begin{theorem}[\textbf{Concept Recoverability via Latent Reconstruction}]
\label{thm:recoverability}
Given Eq.~\eqref{eq:1}, we 
define $z_0 = \arg\min_z \mathcal{F}(y_c, \theta_0, z)$ and $z_e = \arg\min_z \mathcal{F}(y_c, \theta_e, z)$ as the optimal latent codes under the original and erased DMs, respectively. After concept erasure, the alignment for concept $c$ is disrupted as:
\begin{equation}
\mathcal{F}(y_c, \theta_e, z_e) > \mathcal{F}(y_c, \theta_0, z_0) \approx 0.
\end{equation}

Under Assumption~\ref{Assumption 1}, the Hessian matrices $\mathcal{H}_z = \nabla_z^2 \mathcal{F}(y_c, \theta, z)$ and $\mathcal{H}_\theta = \nabla_\theta^2 \mathcal{F}(y_c, \theta, z)$ are invertible, and let $\mathcal{H}_{\theta z} = \nabla_\theta \nabla_z \mathcal{F}(y_c, \theta, z)$ denote the mixed Hessian matrix. By the implicit function theorem, the erased concept can be recovered by fine-tuning the parameters, thereby inducing latent reconstruction. 

Starting from the erased state $(\theta_e, z_e)$ where $\mathcal{F}(y_c, \theta_e, z_e) > 0$, we seek to recover the original alignment $\mathcal{F}(y_c, \theta_0, z_0) \approx 0$. The required parameter shift is:
\begin{equation}
\delta_\theta = \mathcal{H}_\theta^{-1} \cdot \left[\nabla_\theta \mathcal{F}(y_c, \theta_0, z_0) - \nabla_\theta \mathcal{F}(y_c, \theta_e, z_e)\right],
\end{equation}
which induces the latent shift:
\begin{equation}
\delta_z = -\mathcal{H}_z^{-1} \cdot \nabla_{\theta z}^2 \mathcal{F}(y_c, \theta, z) \cdot \delta_\theta,
\end{equation}
such that $\mathcal{F}(y_c, \theta_e + \delta_\theta, z_e + \delta_z) \to \mathcal{F}(y_c, \theta_0, z_0) \approx 0$, thereby recovering the concept.
\end{theorem}

\section{Method}
\label{sec:method}
\subsection{Problem Formulation}
We partition the concept set as $\mathcal{C} = \mathcal{C}_{e} \cup \mathcal{C}_{p}$, where $\mathcal{C}_{e}= \{c_{e}^{1}, \dots, c_{e}^{M}\}$ and $\mathcal{C}_{p}= \{c_{p}^{1}, \dots, c_{p}^{N}\}$ denote the sets of erased and preserved concepts, respectively, with $\mathcal{C}_{e} \cap \mathcal{C}_{p} = \emptyset$.
After concept erasure, concepts in $\mathcal{C}_{e}$ are expected to remain suppressed. From the alignment-function perspective, this process can be formulated as $\mathcal{F}(y_c, \theta_e, z)>\mathcal{F}(y_c, \theta_0, z) \approx 0$, where $\theta_e$ denotes the model parameters after erasure, $\theta_0$ denotes the original pre-trained parameters, and $z$ is the latent code. Our objective is to identify a parameter shift $\Delta\theta$ such that 
$\mathcal{F}(y_c, \theta_e+\Delta\theta, z) \approx \mathcal{F}(y_c, \theta_0, z), \forall c \in \mathcal{C}$. Intuitively, for erased concepts in $\mathcal{C}_{e}$, the parameter shift $\Delta\theta$ aims to restore the model’s suppressed generative capability, effectively reawakening the corresponding concepts. For preserved concepts in $\mathcal{C}_{p}$, the recovery process is required to maintain their original generation behavior, thereby avoiding interference with non-erased semantics.

\subsection{Overview}
An overview of LURE is shown in Fig.~\ref{fig:enter-label}. 
First, we introduce a semantic re-binding mechanism that reconstructs the latent space by aligning the model's denoising predictions with the target concept distributions, re-establishing the severed text-visual associations disrupted by erasure.
When recovering multiple concepts within a shared parameter space, naive joint optimization can lead to gradient conflicts, resulting in feature entanglement across concepts. To address this issue, we impose a gradient orthogonalization constraint to explicitly enforce orthogonality among concept-specific latent feature embeddings. This constraint effectively disentangles optimization directions to mitigate cross-concept interference and enable stable multi-concept recovery.
Second, during inference, we employ LSIS to constrain the sampling trajectory to the high-density manifold of the target concept. By performing posterior consistency verification, LSIS prevents semantic drift and stabilizes the generation process.
Notably, LURE can simultaneously reawaken multiple erased concepts while preserving generative quality and semantic fidelity.

\subsection{Latent Space Reconstruction}
Concept erasure disrupts the implicit mapping between textual concepts and their corresponding latent distributions, rendering certain concepts inaccessible during generation. As discussed earlier, we model the diffusion process as an implicit function jointly parameterized by the textual condition, latent variables, and model parameters. Under this formulation, we recover the erased concepts by reconstructing the latent space to restore the previously disrupted concept–text mapping. To achieve this, we introduce two losses to reconstruct the latent space, a semantic re-binding loss $\mathcal{L}_{\text{Bind}}$ and a gradient field orthogonalization loss $\mathcal{L}_{\text{orth}}$. The formulations of these losses are detailed in the following sections.

\paragraph{Semantic Re-binding Loss.} 
To reconstruct the latent space and reawaken the erased concepts, we consider the $m$-th erased concept $c_m \in \mathcal{C}_e$, where $m \in \{1, \ldots, M\}$. We treat the exemplar set $\mathcal{E}_m = \{x_1, x_2, \ldots, x_S\}$ as observed samples from the target distribution $q(x|c_m)$, where $S$ is the number of exemplar images and each $x$ corresponds to concept $c_m$. We then reconstruct the latent space to restore the original textual–latent mapping disrupted by concept erasure. Specifically, we design a semantic re-binding loss defined as:
\begin{equation}
\label{bind}
\small
    \mathcal{L}^m_{\text{Bind}} = \mathbb{E}_{t, x, \epsilon} \left[ \left\| \epsilon - \epsilon_\theta\left(\sqrt{\bar{\alpha}_t} x + \sqrt{1-\bar{\alpha}_t} \epsilon, t, \tau(p_m)\right) \right\|^2 \right],
\end{equation}
where $p_m$ denotes the text prompt associated with concept $c_m$, and $\tau(\cdot)$ is the text encoder. This term encourages the denoising network to recover accurate noise predictions under the erased textual condition.
From a geometric perspective, this loss guides the gradient flow in latent space toward the manifold regions occupied by $c_m$, effectively ``re-binding" the text token to its corresponding visual representation.

\paragraph{Gradient Field Orthogonalization Loss.} 
Unlike existing methods that focus on reawakening a single erased concept, our framework reconstructs the latent space, which naturally extends to the multi-concept reawaken setting.
However, as established in Theorems~\ref{theorem Concept Entanglement Theory} and \ref{Gradient-Conflict}, jointly reconstructing multiple concepts introduces a \textit{gradient conflict} issue. Specifically, gradients corresponding to different concepts may interfere with one another, which leads to two types of degradation: (i) mutual interference among erased concepts, and (ii) unintended distortion of preserved concepts during recovery. To mitigate this, we introduce a loss function that explicitly encourages a disentangled feature space. This design ensures that reconstructing the latent space for one concept does not adversely affect the latent spaces of other concepts.

Let $\mathbf{v}_{e}^{m} = \epsilon_\theta(z_t, t, y_{c_m})$ represent the latent feature embedding for the $m$-th erased concept $c_m \in \mathcal{C}_e$. Similarly, for any other concept $c_d \in \mathcal{C}$ with $c_d \neq c_m$, we define its latent feature embedding as $\mathbf{v}^{d} = \epsilon_\theta(z_t, t, y_{c_d})$. Then, we define the gradient field orthogonalization loss as:
\begin{equation}
\label{orth}
\small
\mathcal{L}_{\text{orth}}
= \mathbb{E}_{\epsilon, t} \left[
\sum_{m=1}^{M}\sum_{d=1}^{D}
\mathbb{I}[\mathbf{v}_e^m \neq \mathbf{v}^d]\,
\frac{\left|\left\langle \mathbf{v}_{e}^{m}, \mathbf{v}^{d} \right\rangle\right|}
{\left\|\mathbf{v}_{e}^{m}\right\|_2 \left\|\mathbf{v}^{d}\right\|_2 + \xi}
\right],
\end{equation}
where $\mathbb{I}[\cdot]$ is the indicator function and $\xi$ is a small constant for numerical stability.

\begin{table*}[t]
\caption{Quantitative evaluation of ImageNette reawakening across erasure scenarios (ACC \%).}
\label{ImageNette1}
\centering
\resizebox{\textwidth}{!}{%
\begin{tabular}{@{}cc|cccccc|cccccc@{}}
\toprule
\multicolumn{2}{c|}{Reawakened Concepts}&\multicolumn{6}{c|}{English Springer, French Horn, Golf Ball, Parachute}&\multicolumn{6}{c}{English Springer, Golf Ball}\\
\midrule
\multicolumn{1}{c|}{Concepts}&\multicolumn{1}{c|}{SD1.4}&UCE&\textbf{LURE}&RECE&\textbf{LURE}&SPEED&\textbf{LURE}&\multicolumn{1}{|c}{UCE}&\textbf{LURE}&RECE&\textbf{LURE}&SPEED&\textbf{LURE}\\
\midrule
\multicolumn{1}{c|}{Cassette Player}&\multicolumn{1}{c|}{70.6}&75.8&80.6&66.4&89.0&73.2&\multicolumn{1}{c|}{85.8}&74.6&80.8&67.2&84.2&74.2&83.8\\
\multicolumn{1}{c|}{Chain Saw}&\multicolumn{1}{c|}{67.8}&73.8&70.8&63.2&74.2&70.4&\multicolumn{1}{c|}{75.6}&73.4&72.6&77.6&81.2&69.8&72.2\\
\multicolumn{1}{c|}{Church}&\multicolumn{1}{c|}{81.8}&79.2&82.4&77.4&83.0&80.2&\multicolumn{1}{c|}{78.6}&77.8&87.2&76.2&83.4&83.4&81.6\\
\multicolumn{1}{c|}{English Springer}&\multicolumn{1}{c|}{95.2}&\cellcolor[HTML]{FFE8E8}0&\cellcolor[HTML]{FFE8E8}92.0&\cellcolor[HTML]{FFE8E8}0&\cellcolor[HTML]{FFE8E8}79.8&\cellcolor[HTML]{FFE8E8}0.6&\multicolumn{1}{c|}{\cellcolor[HTML]{FFE8E8}95.8}&\cellcolor[HTML]{FFE8E8}0&\cellcolor[HTML]{FFE8E8}84.8&\cellcolor[HTML]{FFE8E8}0&\cellcolor[HTML]{FFE8E8}85.8&\cellcolor[HTML]{FFE8E8}0.8&\cellcolor[HTML]{FFE8E8}97.0\\
\multicolumn{1}{c|}{French Horn}&\multicolumn{1}{c|}{100}&\cellcolor[HTML]{FFE8E8}0&\cellcolor[HTML]{FFE8E8}100&\cellcolor[HTML]{FFE8E8}0&\cellcolor[HTML]{FFE8E8}99.0&\cellcolor[HTML]{FFE8E8}1.2&\multicolumn{1}{c|}{\cellcolor[HTML]{FFE8E8}100}&99.4&99.8&98.8&100&81.6&100\\
\multicolumn{1}{c|}{Garbage Truck}&\multicolumn{1}{c|}{85.0}&88.2&92.2&84.2&89.2&80.4&\multicolumn{1}{c|}{88.6}&86.6&90.6&88.2&88.8&85.2&89.2\\
\multicolumn{1}{c|}{Gas Pump}&\multicolumn{1}{c|}{79.2}&75.8&69.8&79.0&68.2&79.4&\multicolumn{1}{c|}{66.2}&77.4&69.0&76.8&64.2&77.2&67.4\\
\multicolumn{1}{c|}{Golf Ball}&\multicolumn{1}{c|}{95.2}&\cellcolor[HTML]{FFE8E8}0.2&\cellcolor[HTML]{FFE8E8}100&\cellcolor[HTML]{FFE8E8}0&\cellcolor[HTML]{FFE8E8}85.8&\cellcolor[HTML]{FFE8E8}42.2&\multicolumn{1}{c|}{\cellcolor[HTML]{FFE8E8}99.8}&\cellcolor[HTML]{FFE8E8}0.2&\cellcolor[HTML]{FFE8E8}97.4&\cellcolor[HTML]{FFE8E8}0&\cellcolor[HTML]{FFE8E8}93.4&\cellcolor[HTML]{FFE8E8}41.4&\cellcolor[HTML]{FFE8E8}100\\
\multicolumn{1}{c|}{Parachute}&\multicolumn{1}{c|}{96.2}&\cellcolor[HTML]{FFE8E8}0&\cellcolor[HTML]{FFE8E8}95.8&\cellcolor[HTML]{FFE8E8}0&\cellcolor[HTML]{FFE8E8}81.4&\cellcolor[HTML]{FFE8E8}1.4&\multicolumn{1}{c|}{\cellcolor[HTML]{FFE8E8}99.4}&94.0&100&93.0&99.8&95.8&100\\
\multicolumn{1}{c|}{Tench}&\multicolumn{1}{c|}{79.6}&79.6&81.6&77.4&82.8&83.6&\multicolumn{1}{c|}{80.4}&79.6&84.6&78.6&89.4&84.0&87.2\\
\midrule
\multicolumn{1}{c|}{Average} & 85.1 & 47.3 & 86.5 & 44.8 & 83.3 & 51.3& 87.0 & 66.3 & 86.7 & 65.6 & 87.0 & 69.3 & 87.8 \\
\bottomrule
\end{tabular}}
\caption*{\footnotesize \raggedright The \textbf{pink background} indicates the erased concept.}
\vspace{-20pt}
\end{table*}

The loss $\mathcal{L}_{\text{orth}}$ encourages the latent feature embedding associated with each erased concept $c_m$ to be orthogonal to those of all other concepts in $\mathcal{C}$, thereby reducing feature entanglement across concepts. 
As a result, this constraint effectively mitigates cross-concept interference among erased concepts while simultaneously isolating the recovery process to prevent unintended degradation of the preserved concepts.

\begin{figure}[!t]
    \centering
    \includegraphics[width=.85\linewidth]{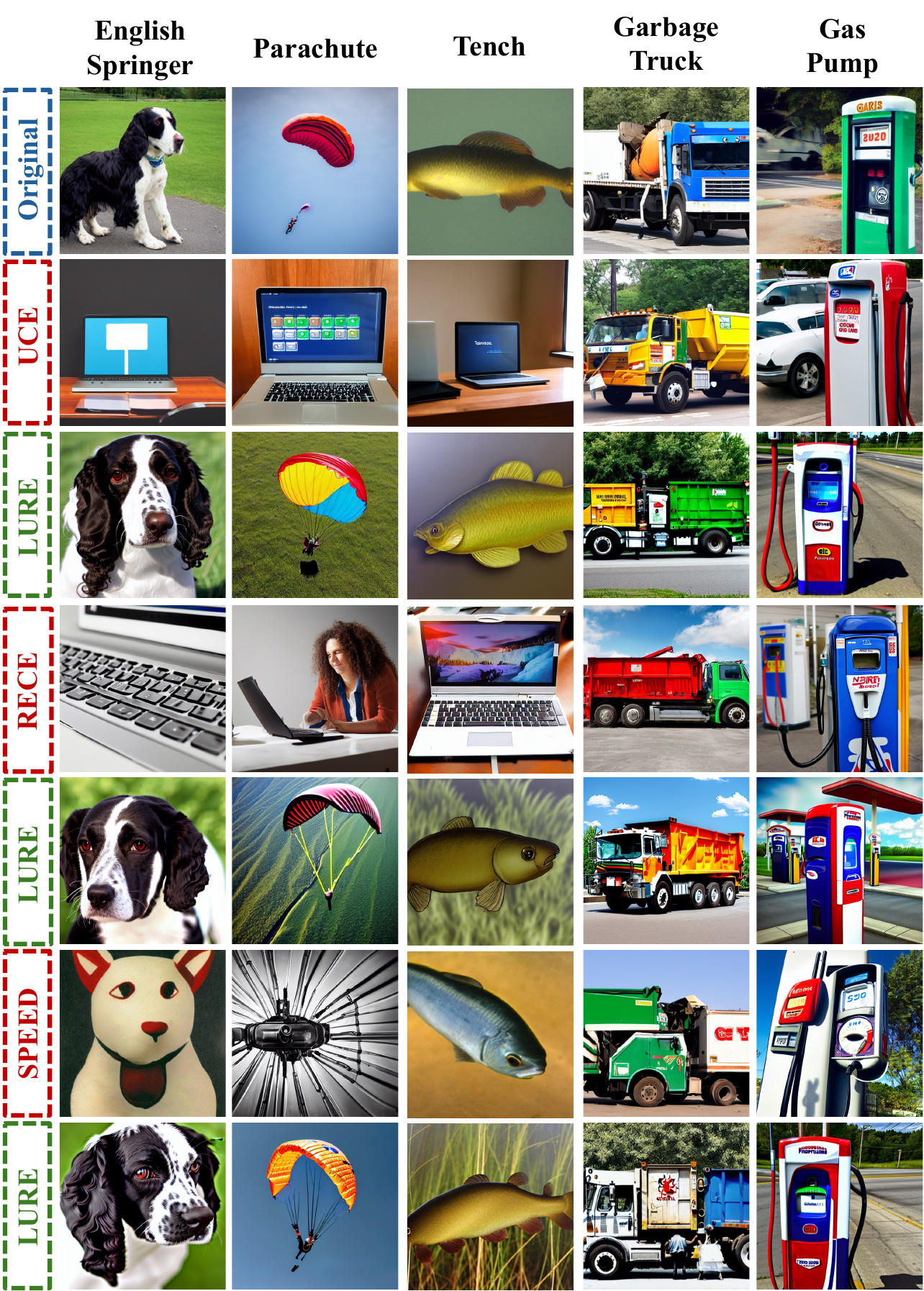}
    \caption{Visual comparison of multi-concept reawakening results (Green box) on objects across UCE, RECE, and SPEED erasure methods (Red box) compared to original SD (Blue box).}
    \label{visual-object}
    \vspace{-10pt}
\end{figure}

\paragraph{Overall Loss Function.} 
The final training loss function combines the semantic re-binding losses for all erased concepts with the gradient field orthogonalization loss. The overall optimization problem is formulated as:
\begin{equation}
\small
    \theta^* = \arg\min_\theta \left( \sum_{m=1}^{M} \mathcal{L}_{\text{Bind}}^m + \lambda \mathcal{L}_{\text{orth}} \right),
\end{equation}
where $\lambda$ is a weighting coefficient that controls the strength of the disentanglement constraint. This formulation encourages the model to reconstruct concept-specific latent representations that satisfy the visual constraints imposed by the exemplar sets, while simultaneously enforcing separation among concept representations in the shared parameter space.

\begin{figure*}[t]
    \centering
    \includegraphics[width=.8\linewidth]{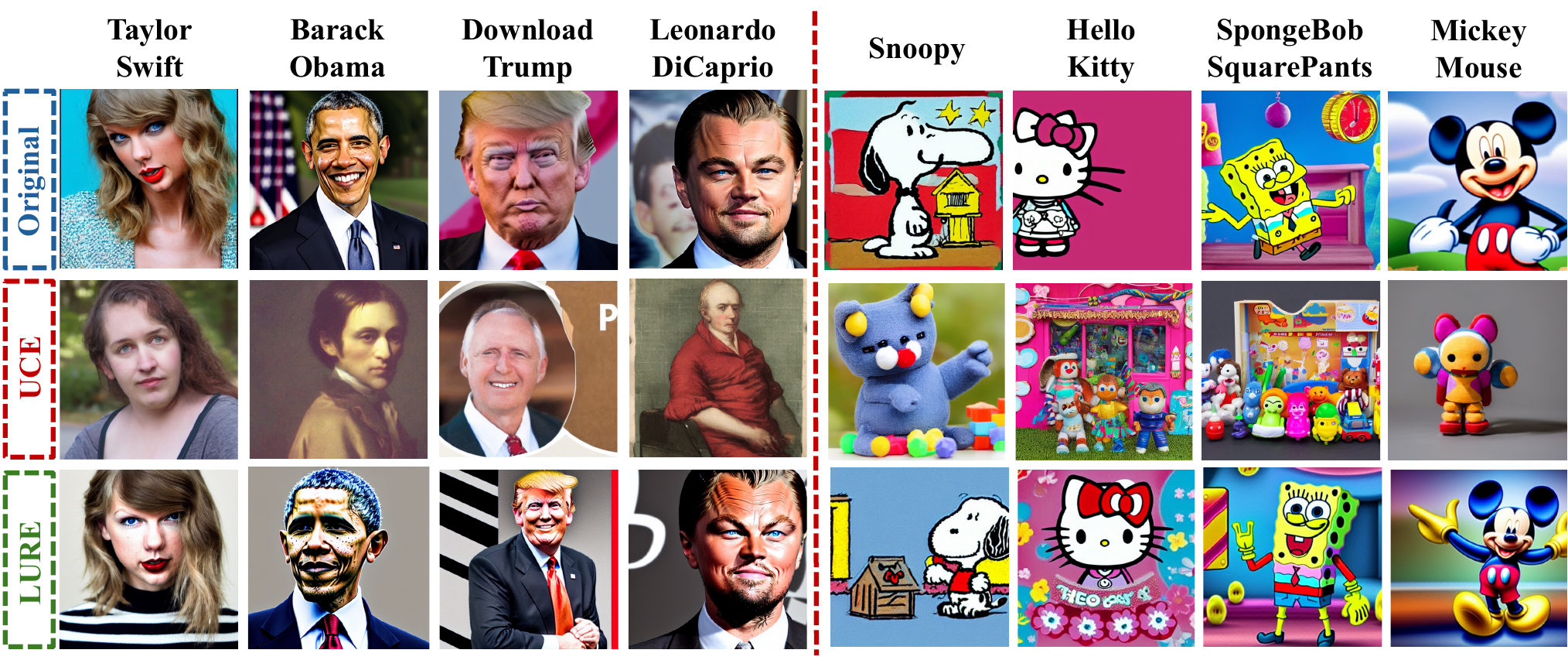}
    \vspace{-5pt}
    \caption{Qualitative comparison of concept reawakening by LURE on Celebrity Identities and Intellectual Property Characters.}
    \vspace{-10pt}
    \label{visual-ip}
\end{figure*}

\subsection{Latent Semantic Identification-Guided Sampling Strategy}
While the above latent-space reconstruction step restores the model’s capacity to generate erased concepts, the generation process may remain unstable during sampling, potentially leading to semantic drift.
To guarantee generation stability and semantic consistency, we introduce an LSIS strategy during the inference phase.

LSIS acts as a sampling-level guidance mechanism that constrains the generation trajectory to generate high-quality, concept-consistent images. Specifically, we observe that latent embeddings at different diffusion timesteps are discriminative with respect to semantic concepts, indicating that they can be used to monitor and regulate concept alignment throughout the generation process. Motivated by this observation, we introduce a lightweight verification module, denoted as $\mathcal{V}_\varphi$. We pre-train $\mathcal{V}_\varphi$ on latent embeddings extracted from real images associated with concepts in $\mathcal{C}$ to capture intrinsic decision boundaries in the latent space. During inference, this module constrains the sampling trajectory by verifying semantic consistency. Given a generated latent code $z$ conditioned on a target concept $c_d$, the verifier $\mathcal{V}_\varphi$ estimates posterior probabilities over the concept set. A generated trajectory is accepted if and only if the target concept is identified as the most probable class, i.e., $\arg\max_{c \in \mathcal{C}} P(c | z) = c_d$.

In this way, LSIS ensures that generated samples remain within the high-density region of the target concept’s latent manifold, effectively preventing semantic drift during sampling. By leveraging both latent space reconstruction and inference-time sampling guidance, LURE achieves stable and faithful reawakening of erased concepts. Moreover, unlike prior works that rely on concept-specific adversarial prompts, our approach naturally enables the simultaneous reawakening of multiple erased concepts within a unified model.

\begin{table*}[!t]
\caption{Perceptual Quality and Semantic Alignment Under Multi-Concept Erasure of English Springer, Tench, and Parachute.}
\vspace{-5pt}
\label{CLIP Score and FID}
\centering
\resizebox{\textwidth}{!}{%
\begin{tabular}{@{}c|ccccccc|cccccc@{}}
\toprule
Metrics         & \multicolumn{7}{c|}{\textbf{CLIP Score $\uparrow$}} & \multicolumn{6}{c}{\textbf{FID Score $\downarrow$}} \\ \midrule
Methods         & Original & UCE & \textbf{LURE} & RECE & \textbf{LURE} & SPEED & \textbf{LURE} & UCE & \textbf{LURE} & RECE & \textbf{LURE} & SPEED & \textbf{LURE} \\ \midrule
English Springer & 33.33 & 14.56 & 32.89 & 14.51 & 32.84 & 24.38 & 32.90 & 244.21 & 32.55 & 244.76 & 33.99 & 231.18 & 46.32 \\
Parachute       & 32.00 & 17.13 & 32.85 & 17.78 & 27.52 & 24.75 & 33.16 & 268.56 & 33.32 & 301.81 & 178.06 & 280.17 & 41.12 \\
Tench           & 32.87 & 17.94 & 30.42 & 17.28 & 32.74 & 31.12 & 34.01 & 243.26 & 73.21 & 246.27 & 46.69 & 30.34 & 23.61 \\ \midrule
Average         & 32.73 & 16.54 & 32.05 & 16.52 & 31.03 & 26.75 & 33.36 & 252.01 & 46.36 & 264.28 & 86.25 & 180.56 & 37.02 \\ \bottomrule
\end{tabular}%
}
\vspace{-10pt}
\end{table*}

\begin{table}[!t]
\caption{Combined CLIP Score for celebrity identities and intellectual property characters.}
\vspace{-8pt}
\label{CLIP-celebrity-IP}
\resizebox{0.95\columnwidth}{!}{%
\begin{tabular}{@{}c|ccccc@{}}
\toprule
\multirow{2}{*}{Concepts} & \multicolumn{5}{c}{\textbf{Celebrity Identities}}                                                                             \\ \cmidrule(l){2-6} 
                                   & \begin{tabular}[c]{@{}c@{}}Taylor\\ Swift\end{tabular} & \begin{tabular}[c]{@{}c@{}}Barack\\ Obama\end{tabular} & \begin{tabular}[c]{@{}c@{}}Donald\\ Trump\end{tabular} & \begin{tabular}[c]{@{}c@{}}Leonardo\\ DiCaprio\end{tabular}      & Average \\ \midrule
Original                           & 29.61                                                  & 29.79                                                  & 28.56                                                  & 33.01                                                            & 30.24   \\\midrule
UCE                                & 20.54                                                  & 21.26                                                  & 21.25                                                  & 18.89                                                            & 20.48   \\
\textbf{LURE}                             & 29.33                                                  & 29.87                                                  & 28.80                                                  & 31.80                                                            & 29.95   \\\midrule
RECE                               & 20.54                                                  & 21.26                                                  & 21.25                                                  & 18.89                                                            & 19.24   \\
\textbf{LURE}                              & 30.01                                                  & 29.99                                                  & 27.96                                                  & 32.20                                                            & 30.07   \\\midrule
SPEED                              & 23.97                                                  & 23.19                                                  & 22.48                                                  & 22.37                                                            & 23.00   \\
\textbf{LURE}                              & 30.11                                                  & 29.84                                                  & 28.70                                                  & 34.11                                                            & 30.69   \\ \midrule
\multirow{2}{*}{Concepts}          & \multicolumn{5}{c}{\textbf{Intellectual Property Characters}}                                                                                                                                                                                         \\ \cmidrule(l){2-6} 
                                   & Snoopy                                                 & \begin{tabular}[c]{@{}c@{}}Hello \\ Kitty\end{tabular} & \begin{tabular}[c]{@{}c@{}}Mickey\\ Mouse\end{tabular} & \begin{tabular}[c]{@{}c@{}}SpongeBob \\ SquarePants\end{tabular} & Average \\ \midrule
Original                           & 33.45                                                  & 32.84                                                  & 32.73                                                  & 31.03                                                            & 32.51   \\ \midrule
UCE                                & 19.43                                                  & 19.06                                                  & 18.90                                                  & 19.58                                                            & 19.24   \\
\textbf{LURE}                              & 32.45                                                  & 32.32                                                  & 32.21                                                  & 30.90                                                            & 31.97   \\ \midrule
RECE                               & 19.43                                                  & 19.06                                                  & 18.89                                                  & 19.58                                                            & 19.24   \\
\textbf{LURE}                              & 31.66                                                  & 31.44                                                  & 31.62                                                  & 28.98                                                            & 30.91   \\ \midrule
SPEED                              & 22.85                                                  & 25.79                                                  & 23.39                                                  & 26.36                                                            & 24.60   \\
\textbf{LURE}                              & 33.86                                                  & 31.88                                                  & 32.58                                                  & 31.01                                                            & 32.33   \\  \bottomrule
\end{tabular}%
}
\vspace{-10pt}
\end{table}

\section{Experiments}
\subsection{Experimental Setup}
\textbf{Dataset and Concept Selection.} 
To comprehensively validate our proposed method, we conduct experiments across four representative reawakening tasks: (i) \textit{\textbf{General Visual Objects.}} We select ten representative classes from ImageNette~\citep{howard2019imagenette} to benchmark semantic recovery. (ii) \textit{\textbf{Intellectual Property Concepts.}} These concepts correspond to copyrighted fictional characters. (iii) \textit{\textbf{Celebrity Identities.}} This category includes real-person identity concepts. (iv) \textit{\textbf{Unsafe Content Categories.}} This setting covers safety-critical concepts, including nudity, blood, and weapons.

\noindent \textbf{Erasure Setup.} We evaluate the performance of LURE against three advanced concept erasure methods. UCE performs one-step tuning on cross-attention layers. RECE achieves concept erasure by modifying cross-attention matrices, iteratively deriving and removing concept embeddings, and incorporating regularization to minimize similarity to the original concept. SPEED realizes concept erasure through the null-space constrained parameter editing, supplemented by multiple prior knowledge refinement strategies.
These methods follow different concept erasure paradigms, which enable a comprehensive evaluation of the effectiveness and robustness of our approach.

\begin{table}[!t]
\caption{Semantic alignment (CLIP Score) for unsafe content concepts across three erasure methods.}
\vspace{-8pt}
\label{Unsafe}
\centering
\resizebox{0.75\columnwidth}{!}{%
\begin{tabular}{@{}c|cccc@{}}
\toprule
Concepts & Blood & Nudity & Weapon & Average \\ \midrule
Original & 26.95 & 30.98 & 28.55 & 28.83 \\ \midrule
UCE      & 24.68 & 21.50 & 20.97 & 22.38 \\
\textbf{LURE}   & 26.74 & 30.94 & 28.56 & 28.74 \\ \midrule
RECE     & 22.20 & 20.59 & 20.16 & 20.98 \\
\textbf{LURE}    & 26.72 & 29.60 & 27.70 & 28.01 \\ \midrule
SPEED    & 23.90 & 24.07 & 24.07 & 23.69 \\
\textbf{LURE}   & 27.14 & 31.96 & 28.17 & 29.09 \\
\bottomrule
\end{tabular}%
}
\vspace{-10pt}
\end{table}

\begin{table}[]
\centering
\caption{Ablation study of different modules.}
\vspace{-8pt}
\label{tab:ablation_object}
\resizebox{\columnwidth}{!}{%
\begin{tabular}{@{}c|c|c|cccc@{}}
\toprule
Concept                                                                     & Metric & Original & \begin{tabular}[c]{@{}c@{}}Joint\\ Tuning\end{tabular} & \begin{tabular}[c]{@{}c@{}}w/o Latent\\ Space Recon.\end{tabular} & \begin{tabular}[c]{@{}c@{}}w/o\\ LSIS\end{tabular} & \begin{tabular}[c]{@{}c@{}}LURE\\ (Ours)\end{tabular} \\ \midrule
\multirow{3}{*}{\begin{tabular}[c]{@{}c@{}}English\\ Springer\end{tabular}} & ACC$\uparrow$    & 95.2     & 9.6                                                    & 91.2                                                              & 89.6                                               & 92.0                                                  \\
                                                                            & CLIP $\uparrow$   & 33.33    & 19.6                                                   & 31.33                                                             & 32.71                                              & 32.89                                                 \\
                                                                            & FID $\downarrow$    & -        & 212.75                                                 & 51.38                                                             & 34.50                                              & 32.55                                                 \\ \midrule
\multirow{3}{*}{Parachute}                                                  & ACC $\uparrow$    & 96.2     & 69.4                                                   & 77.4                                                              & 78.2                                               & 95.8                                                  \\
                                                                            & CLIP $\uparrow$   & 32.00    & 29.67                                                  & 31.78                                                             & 30.38                                              & 32.65                                                 \\
                                                                            & FID $\downarrow$    & -        & 147.54                                                 & 42.31                                                             & 67.69                                              & 33.32                                                 \\ \midrule
\multirow{3}{*}{Tench}                                                      & ACC $\uparrow$    & 79.6     & 13.2                                                   & 38.2                                                              & 36.2                                               & 69.6                                                  \\
                                                                            & CLIP $\uparrow$   & 32.87    & 27.46                                                  & 28.36                                                             & 26.34                                              & 30.92                                                 \\
                                                                            & FID $\downarrow$    & -        & 94.37                                                  & 108.46                                                            & 122.99                                             & 53.21                                                 \\ \bottomrule
\end{tabular}%
}
\vspace{-15pt}
\end{table}

\noindent \textbf{Experimental Setting.} 
For each erased concept, we construct a minimal exemplar set by randomly sampling three images. The erased model is then fine-tuned for 1,000 steps with a learning rate of $5 \times 10^{-6}$ and a batch size of 4. During training, the text encoder is frozen, and only the U-Net of the DM is optimized. All experiments are conducted using SD v1.4 on a single NVIDIA RTX 4090 GPU.

\noindent \textbf{Evaluation Metrics.}
For each reawakened concept, we generate 500 images using the corresponding text prompts to ensure a fair evaluation. We employ three complementary metrics. Classification Accuracy (ACC) using ResNet-50 pre-trained on ImageNet measures semantic correctness. Fréchet Inception Distance (FID) assesses the distributional quality and visual fidelity of the generated images. CLIP Score evaluates text-image semantic alignment.

\subsection{Multi-Concept Recovery Performance}
Most existing evaluation protocols focus on single-concept reawakening. Latent space reconstruction enables consideration of the challenging multi-concept scenario. Therefore, we evaluate our method across three different reawaken tasks.

\paragraph{General Visual Objects.}
Fig.~\ref{visual-object} and Tab.~\ref{ImageNette1} show that LURE effectively reawakens erased concepts across various erasure methods, while preserving the generation capability of retained concepts. 
Further validation in Tab.~\ref{CLIP Score and FID} confirms that while erasure baselines successfully suppress target concepts, LURE consistently restores both semantic alignment and distributional fidelity.
Specifically, the recovery in CLIP scores confirms that $\mathcal{L}^m_{\text{Bind}}$ effectively rectifies the latent gradient flow, realigning the sampling trajectory with the target concept manifold. Meanwhile, the high classification accuracy verifies that $\mathcal{L}_{\text{orth}}$ achieves effective feature disentanglement in the parameter space by ensuring that optimization directions for different concepts remain non-interfering. Finally, the improved FID scores indicate that LSIS-constrained inference successfully restricts sampling to high-density posterior regions, thereby filtering out off-manifold samples that cause semantic drift.

\paragraph{Intellectual Property Concepts.}
Fig.~\ref{visual-ip} and Tab.~\ref{CLIP-celebrity-IP} demonstrate the effectiveness of LURE on the intellectual property concept reawakening task. The reawakened models are able to generate recognizable intellectual property characters and celebrity identities with high image quality and clear identity characteristics. Notably, even for this task, which requires the recovery of rich and fine-grained details, LURE consistently produces high-quality images.

\paragraph{Unsafe Content Categories.}
Beyond general visual objects and intellectual property concept reawakening tasks, which require the recovery of fine-grained details, we further evaluate LURE on reawakening concepts from the unsafe content category. This task requires the model to correctly interpret and recover suppressed safety-related semantics.
As shown in Table~\ref{Unsafe}, LURE successfully bypasses the safety filters imposed by UCE, RECE, and SPEED, which enables the DM to generate images associated with concepts such as blood, nudity, and weapons. These results indicate that LURE effectively reconstructs the latent space and restores the underlying semantic representations that are suppressed by existing safety-oriented erasure mechanisms.

\subsection{Ablation Study}
We conduct an ablation study to verify the effectiveness of our framework's modules, as shown in Table~\ref{tab:ablation_object}. In particular, the performance degradation observed with naive fine-tuning is attributed to severe gradient conflicts within the shared parameter space, where feature entanglement prevents effective re-binding of severed text-visual associations. In contrast, the instability observed when LSIS is removed arises from the absence of posterior density verification during sampling. Without this constraint, the sampling trajectory can drift semantically away from the high-density regions of the target concept manifold, thereby compromising generation fidelity.

\section{Conclusion}
In this work, we expose vulnerabilities in concept erasure via LURE, a unified framework for multi-concept reawakening. By modeling generation as an implicit function of textual conditions, model parameters, and latent states, we theoretically show that perturbing these factors can reawaken erased concepts.
Leveraging this, LURE reconstructs the latent space via Semantic Re-binding to restore disrupted text-visual associations, while the Gradient Field Orthogonalization mechanism enforces orthogonality among latent feature embeddings, thereby disentangling conflicting representations and enabling interference-free joint optimization. Furthermore, the LSIS strategy provides essential post hoc verification, ensuring that generated samples remain within the target concept manifolds and preventing semantic drift. 
Comprehensive evaluations validate the effectiveness of LURE in achieving high-fidelity reawakening of multiple concepts. Taken together, our findings underscore that robust safety requires fundamentally eliminating latent generative knowledge, rather than simply suppressing its expression.

\section*{Acknowledgments}
This work was supported by the National Natural Science Foundation of China under Grant 62271335, 62502416, U25A20430, Sichuan Science and Technology Program under Grant 2025ZNSFSC0470, and the Research Grants Council (Grant No: 15224124), Hong Kong SAR, China.

\bibliographystyle{named}
\bibliography{ijcai26}

\end{document}